\documentclass[runningheads]{llncs}
\usepackage{graphicx}
\usepackage{xcolor}
\usepackage{multirow}
\usepackage{float}

\begin{document}

\title{Engineering Features to Improve Pass Prediction in Soccer Simulation 2D Games}
\titlerunning{Engineering Features to Improve Pass Prediction}
%
\author{Nader Zare\inst{1} \and
Mahtab Sarvmaili\inst{1} \and
Aref Sayareh\inst{3} \and
Omid Amini\inst{4} \and
Stan Matwin\inst{1,2} \and
Amilcar Soares\inst{5}}
\authorrunning{N. Zare et al.}
%
\institute{Institute for Big Data Analytics, Dalhousie University, Halifax, Canada \\
\and
Institute for Computer Science, Polish Academy of Sciences, Warsaw
\and
Shiraz University, Shiraz, Iran\\
\and
Qom University of Technology, Qom, Iran\\
\and
Memorial University of Newfoundland, St. John's, Canada\\
\email{\{nader.zare, mahtab.sarvmaili\}@dal.ca, arefsayareh@gmail.com, omidamini360@gmail.com, stan@cs.dal.ca, amilcarsj@mun.ca}
}
\maketitle              

\begin{abstract}

Soccer Simulation 2D (SS2D) is a simulation of a real soccer game in two dimensions. 
In soccer, passing behavior is an essential action for keeping the ball in possession of our team and creating goal opportunities. 
Similarly, for SS2D, predicting the passing behaviors of both opponents and our teammates helps manage resources and score more goals. 
Therefore, in this research, we have tried to address the modeling of passing behavior of soccer 2D players using Deep Neural Networks (DNN) and Random Forest (RF). 
We propose an embedded data extraction module that can record the decision-making of agents in an online format. 
Afterward, we apply four data sorting techniques for training data preparation.
After, we evaluate the trained models' performance playing against 6 top teams of RoboCup 2019 that have distinctive playing strategies. Finally, we examine the importance of different feature groups on the prediction of a passing strategy. 
All results in each step of this work prove our suggested methodology's effectiveness and improve the performance of the pass prediction in Soccer Simulation 2D games ranging from 5\% (e.g., playing against the same team) to 10\% (e.g., playing against Robocup top teams). 


\vspace{2em}
\textbf{NOTICE:} This is an accepted article to be published by Lecture Notes in Artificial Intelligence (LNCS/LNAI) series by Springer-Verlag in the proceedings of the RoboCup International Symposium.
\end{abstract}
\section{Introduction}

Soccer is the world's most popular sport where two teams of eleven players play against each other, and the team with the highest number of goals wins the game. 
Shooting, dribbling, and passing are examples of possible actions that can lead the ball to the goal. 
Possessing the ball for a more extended period not only reduces the chance of losing the game to the opposing team but may also create more opportunities to shoot at the opponent's goal. 
The team with a smart passing strategy dictates the play, saves energy, and makes the best use of their resources (e.g., player's stamina).

Due to technological development, the idea of robotic soccer players was introduced as a new research area in 1992 \cite{mackworth1993seeing}. 
Since then, the RoboCup\footnote{www.robocup.org} has been known as an international competition for promoting ideas in A.I. and robotics. 
Within this competition, the 2D soccer simulation league works as an abstraction of real soccer games into a two-dimensional environment.

The RoboCup Soccer Simulation Server (RCSSServer) is responsible for simulating the environment and the connection between the game elements. 
For this simulation each player (agent) is an individual program that communicates with the server to send actions and to receive its observations.

As in real soccer, passing behavior in a Soccer Simulation 2D (SS2D) game plays a critical role in increasing the chance of winning.
The prediction of agents' passing behavior has the advantages of improving a team's passing strategy, increasing the accuracy of passing actions, enhancing agents' decision-making, managing agents' stamina, and enhancing an agent's unmark behavior.
In this work, we propose an embedded Data Extractor module that collects events of an SS2D game and creates features that can be used to train models to enhance the passing actions in a game.

This module generates data containing features of the ball, our players, and opponent players.
We also propose three sorting methods along with two modes of placing kicker's features as the first element of each data instance to create.
We trained a Deep Neural Network (DNN) and Random Forest (RF) models on a generated dataset with 3,000 games to predict our teammates' passing behavior to validate our module. 
Our experimental results show that these sorting algorithms, along with suggested feature groups, improve the team's passing prediction accuracy. 
To assess the quality of the suggested feature groups, we investigated the robustness of these feature groups in the face of changes in the opponent team.

The rest of this paper is organized as follows. Section \ref{sec:relatedworks} presents some related works in the area of soccer simulation 2D. In Section \ref{sec:environment}, we provide definitions of the environment and some detail about soccer simulation. Section \ref{sec:dataextractor} explains the data gathering step. Section \ref{sec:experimental} shows the experimental results of our model. Finally, Section \ref{sec:conclusion} concludes the work and also discusses future works.

\section{Related Works}
\label{sec:relatedworks}

The prediction of players' behavior is an active research topic in robotic soccer. 
The work of \cite{nakashima2015kick} tries to predict the tactical strategy of opponent by analyzing the kicking distribution of agents from the records of the game.
Similarly, \cite{fukushima2019similarity} has evaluated the kicking distribution and action trajectories of players from records of games to analyze teams' offensive strategy. 
The positional features of the players and the ball extracted from the records of the game have been a source of data for the behavioral analysis of players, such as the proposed methods in these papers \cite{michael2017analysing,asali2016using,suzuki2019use} extracted 46 positional features of players and ball in the field for trajectory analysis, formation detection and assessment of games' state respectively, without any further investigation of the effect of other features groups. 
All these works had applied machine learning algorithms to the records of the game that were extracted when the game was finished; therefore the observation of agents and the process of decision making remains unclear. 

Unlike all previous approaches, we propose an online data extractor inserted in the players of a team, which continuously records the agent's action from the environment. 
In this case, not only do we have access to the decision of agents but also, we can record the live observations of the agent from the environment. 
Additionally, we have introduced different feature groups and sorting methods. 
Our idea is to evaluate the impact of these feature groups on the prediction of passing behavior. 

To the best of our knowledge, implementing the data generator module inside the players and performing the online recording of the game's information is done for the first time in Soccer Simulation 2D research. 
None of the previous researches have explored the significance of other positional feature groups rather than Cartesian positions of elements in the field. 
Alternatively, investigate the robustness of a trained model against changes in the opponent team has never been explored in the literature.

\section{The environment}
\label{sec:environment}
\begin{figure}
    \centering
    \includegraphics[scale=0.58,trim={0.64cm 0 0.64cm 0},clip]{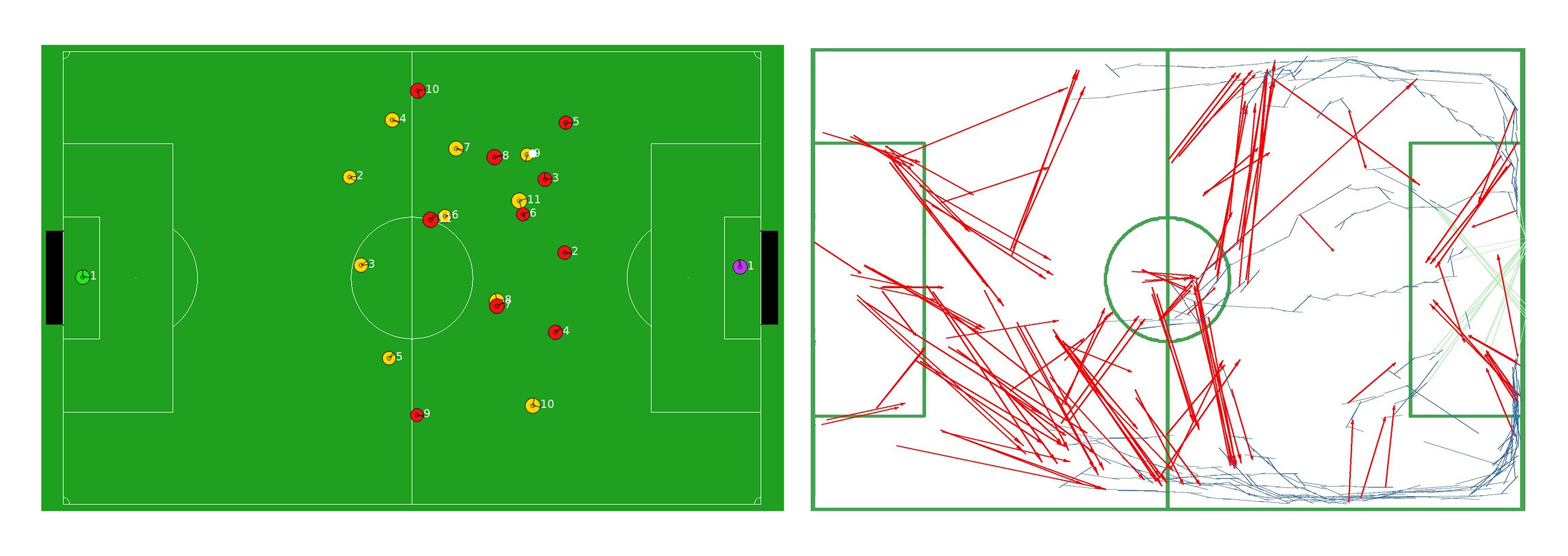} \\
    \caption{Left: An overview of Soccer Simulation 2D Environment. Right:Visualization of all passes, dribbles and shoots between two agents from the left team.}
    \label{fig:2denv}
\end{figure}

Soccer Simulation 2D (SS2D) League is part of the RoboCup \cite{kitano1997robocup} research project and presents a dynamic, continuous, real-time, incomplete, and distributed environment. 
Like real soccer, the teams of eleven players compete against each other in a two-dimensional field set up on a computer server. 
All the players are mapped to this two-dimensional space, and they are represented by circles (Figure \ref{fig:2denv}).

Each team consists of 12 agents (11 players and one coach), and each agent is executed as an independent program. 
A game in this league consists of 6,000  cycles, and one cycle is represented in 0.1 seconds.

In a SS2D game, the players of a team are created before a match starts.
The server will randomly generate 18 players with the attributes: $maximum\ speed$, $decay$, $size$, $effort\ max$, $effort\ min$, $kickable\ area\ size$, $kick\ power$, $margin$, $dash\ rate$. Afterwards, the server sends all 18 different players to the teams' coach, who has the task of selecting 11 players to start a game.

At the beginning of each cycle, the server sends distinct information to each agent based on its angle view in the previous cycle. 
At the end of each cycle, players send their actions along with their parameters to the server. 
After, the server adds a noise to the received actions from all players and it updates the state of game, moving to the next cycle at the end of this step. 
The result of repeated matches between two teams can be different every time a game is played since the noise values are randomly generated in each game, even if they use the same algorithms and parameters in those matches.
\vspace{-5mm}
\setlength{\abovecaptionskip}{0pt}
\setlength{\belowcaptionskip}{-30pt}

\subsection{The Helios Base}
\vspace{0mm}
Most teams that play in the SS2D league use one of the popular released base codes \cite{akiyama2013helios}\cite{prokopenko2019gliders2d}\cite{kok2003uva}\cite{cyrus19}. 
The HELIOS base \cite{akiyama2013helios} is the most popular base code among the teams. 
It consists of a debugging monitor, a formation editor, and a sample team (known as \emph{Agent2D}) that can connect to the server and plays a full game against other teams. 
Some of the fundamental behaviors such as passing, dribbling, and shooting have been developed in this sample team.
In this work, we are giving an overview of the \emph{Agent2D} Base but full details and documentation are available in \cite{akiyama2013helios}.

The \emph{Chain Action} is a decision-making module that has been developed for the player who holds the ball (i.e., ball owner or ball holder) in \emph{Agent2D} base. 
The module uses the Breadth-First Search algorithm (BFS) \cite{mugnier1992conceptual} to find the best action for the ball holder. 
First, it creates a decision tree where the root node is the current state, and the actions are branches from this state. 
In the tree, each edge is a possible action selected (e.g., pass, shoot, dribble, etc), and each node is a state (e.g., the position of the ball and players). Second, the \textit{Field Evaluator} evaluates all nodes by using the position of the ball in that node. Finally, it selects the action that leads to the node with highest value of this metric. Figure \ref{fig:2denv} presents all selected action by players' of left team in a game.

The agent who has the ball can use the predictive model before receiving the ball to update his neck angle and to see the target player to enhance its pass accuracy. 

This agent can also use the predicting model to deeply search the action tree, aiming to find the best chain of actions in the field. 
Finally, agents in the SS2D environment, like in real soccer games, should move to a position to receive the ball. 
In the SS2D league, each agent has limited stamina (i.e., players have a limited energy capacity in the game), and they can use this energy for kicking and moving. 
Efficient management of stamina is required, so all players can't go to a position to be the pass target. Therefore, deciding which player is the most likely to receive an accurate pass helps manage their available stamina.

\section{The Data Extractor Module}
\label{sec:dataextractor}

In this work, we developed a \emph{Data Extractor} Module to create training data to feed machine learning models in an SS2D game. Our module was designed in the Helios base sample team (Agent2D) and collects the events of the game, transforming these events into training data for machine learning models. 

\begin{figure}
    \centering
    \includegraphics[scale=0.20]{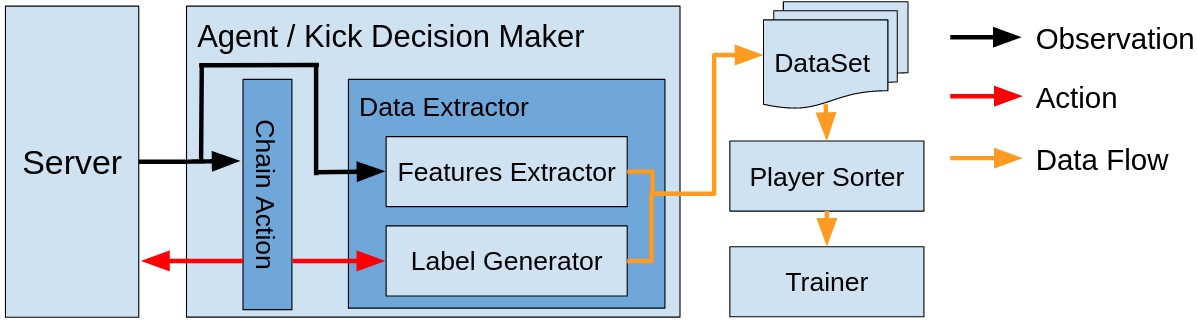} \\
    \caption{Overview of the Data Extractor module.}
    \label{fig:dataflow}
\end{figure}

As mentioned previously, each agent receives its observation data from the server. If the agent is the ball owner, it feeds the observed information to the Chain Action and Data Extractor modules. 
The \emph{Chain Action} module finds the best action and sends it to the \emph{Data Extractor} module, only if the selected action is a pass.
The \emph{Data Extractor} Module has two sub-modules named \emph{Feature Extractor} and \emph{Label Generator}.
The \emph{Data Extractor} module gets the observation and the best action, and delivers the observation and action to the \emph{Features Extractor} and \emph{Label Generator}. 
The \emph{Features Extractor} extracts the list of features shown in Section \ref{sec:features} from the observation.
The \emph{Label Generator} assigns one label $"Unum(Uniform\,Number)"$ on the generated features. 
Finally, the \emph{Data Extractor} saves the data to be used for training purposes.
After completion of game, the generated dataset is sent to the \emph{Sorting} module. 
This module reads the data features and sorts them based on six methods that are explained in \ref{subsec:sorting}. 
Eventually, the external \emph{Trainer} Module receives sorted data and employs it for training the model.  
The structure of the agent and its processing modules are presented in the Figure \ref{fig:dataflow}. 
Next, we discus the full procedure in details. 

\vspace{-5mm}
\setlength{\abovecaptionskip}{0pt}
\setlength{\belowcaptionskip}{-30pt}

\subsection{Features and Values}
\label{sec:features}
For each cycle that one of our agents is the ball holder and the selected action is a pass, the agent creates a new data instance. 
This instance contains 12 features for the ball, 42 for each one of our players ($42\times 11$), and 24 for each of the opponent's players ($24 \times 11$), in total $738$ features.
We used two labels for each data instance: Index Number and Uniform Number (Unum). 
The Uniform Number is the unique number of a target agent (who may receive the ball) in the game. 
Differently from Unum, the Index Number refers to the index of a target agent in the sorted data. 
The Index Number is assigned to the player in the \emph{Player Sorter} module.

The list of extracted feature are broken down into nine feature subgroups that are measured for the ball, our agents, or the opponents' agents. Since at each time step, the ball holder is responsible for generating a data instance, the module creates all of these features for all agents on the field. In the following, we will discuss them in more detail.


\textbf{Position.} This feature group measures the position of an object in the field by using 4 parameters $X$, $Y$, $R$, and $T$. $X$ and $Y$ are the vector's coordinates in Cartesian system, $R$ and $T$ are the vector's coordinate in Polar system.

\textbf{Kicker.} This group measures the position of an object in the field with respect to the position of the ball holder. It contains four positional features similar to the Position feature group.

\textbf{Velocity.} This group measures the velocity of an object in the field by using positional parameters. It contains four positional features similar to the Position feature group.

\textbf{Body.} It measures the players' body angle with respect to the direct line between the center of two goals.

\textbf{Team.} This group contains two sets of parameters, Uniform number and "Is Kicker". Uniform number presents the unique number of players and "Is kicker" shows that the player is ball holder or not.

\textbf{Player Type.} This feature group contains nine values for player type.

\textbf{Top k-th riskiest opponents.}  This feature group is measured for all of our players and helps to identify the top k riskiest opponents on the way of a pass to another teammate. 
The risk of an opponent is calculated by the difference between two angles: 1) angle of the ball to the opponent, 2) angle of the ball to the teammate.

\begin{figure}
    \centering
    \includegraphics[scale=0.30]{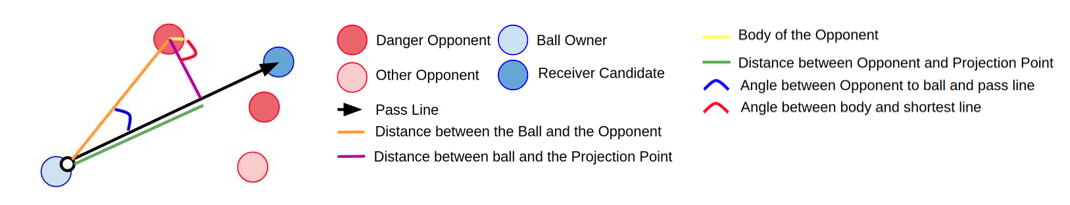}
    \caption{Illustration of calculating "Top Kth riskiest" feature group}
    \label{fig:passcut}
\end{figure}

The ball holder draws a direct line to all of the opponents and calculates it as the angle to the line between the ball and the target teammate. 
After identifying the two riskiest opponents, the agent will measure 
1) the distance between the opponent and the ball, 
2) the distance between the opponent and the direct pass line, 
3) the angle of the line between the ball and the opponent, and the direct pass line,
4) the angle between the opponent body and the perpendicular line to the direct pass line, and 
5) the distance between the projection point and the ball holder. 
Figure \ref{fig:passcut} illustrates the discussed features in the top k riskiest opponents' agents.

\textbf{Top k-th nearest opponents.} This feature group shows relative information of the k-th nearest opponents to the teammate. It is identified by three elements, \textit{distance}, \textit{angle}, and \textit{body angle} of the opponent player to ours.

\textbf{Goal.} This feature group indicates the distance and the angle between teammates and the center of the opponent's goal.

The Position, Kicker and the Velocity groups are measured for all objects in the field. the Body, Team and the Player Type are measured for all players. The rest of feature groups are only measured for our players.

\vspace{-5mm}
\setlength{\abovecaptionskip}{0pt}
\setlength{\belowcaptionskip}{-15pt}

\subsection{Sorting}
\label{subsec:sorting}

The \emph{Sorting} module is an essential component in the data preparation step since it organizes the input features and creates the training dataset.
We proposed 4 distinct methods for the \emph{sorting} sub-module.

\textbf{Uniform Number Sorting.} In this approach the sorting module arranges the data by uniform number. The data instance will be structured by: features of the Ball, features of Teammates sorted by Uniform number, and features of Opponents sorted by Uniform number.

\textbf{X Sorting} For this method, the sorting module arranges the data based on the x coordinate, and the data will be structured by: features of the Ball, features of Teammates sorted, and features of Opponents sorted by X respectively.

\textbf{Field Evaluator Sorting} This method sorts players based on an evaluation of their positions from the Field Evaluator sub module in the \emph{Chain Action}\cite{akiyama2013helios}. The structure of the data will be as features of the Ball, features of Teammates, features of Opponents sorted by the evaluation criteria, respectively.

\textbf{Kicker be First} This is a binary attribute that pushes the features of the ball holder as the first element of data. 

After sorting the data, this module assigns a label to them which is based on the index of ball receiver. 
When the sorter module puts the information of players in one data instance, it finds the index of ball receiver in the sorted features. For example, the features of the ball receiver is the fifth elements of data. 
In this case the label of this data record will be five.
Therefore applying three sorting methods and changing the \textbf{Kicker be First} attribute (i.e., true or false), we can generate six different datasets with a different order of the input data.

\vspace{-5mm}
\section{Experimental Results}
\label{sec:experimental}
\subsection{Training Data Set}

To obtain the training data \footnote{The dataset and source of this project for reproducing the results are available in "https://github.com/Cyrus2D/Agent2D-DataExtractor" }, we enabled a "full state option" in the server that requires the server to dispense the exact observations of the environment to players. This feature improves the reproducibility of our model. For this work, we set the parameter of K (employed for the "Top Kth riskiest opponents" and "Top Kth nearest opponents") equal to two.

To create the dataset, we inserted the \emph{Data Extractor} Module in Agent2D base, and we played against the same source code for 3,000 games (i.e. both teams are using the same source code, but only one of them has \emph{Data Extractor} Module) and 1,361,126 data instances were generated. 

\begin{figure}[H]
    \centering
    \includegraphics[scale=0.33]{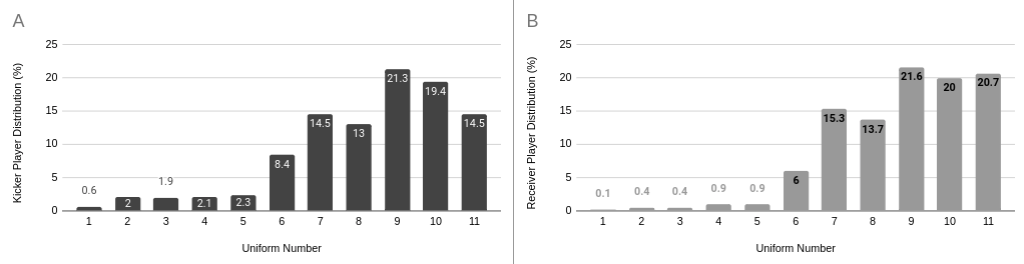} 
    \caption{The probability distribution of being the Kicker or the Receiver for all players}
    \label{fig:dist}
\end{figure}

We split the data into two subsets of training with $1,088,237$ instances and tested with $272,889$ instances. The distribution of acting as kicker players and being the ball's receiver is shown in Figure \ref{fig:dist}.
To sort the train and test data, we applied the three sorting methods (discussed in \ref{subsec:sorting}) in two modes where the \textit{Kicker be First} attribute is $True$ or $False$.

\vspace{-5mm}
\setlength{\abovecaptionskip}{0pt}
\setlength{\belowcaptionskip}{-15pt}


\subsection{Early results}
For the prediction model, we employed a Deep Neural Network (DNN) and a Random Forest (RF).

The DNN model includes 6 layers of $1024$, $512$, $256$, $64$, $32$, and $11$ neurons. 
The activation function of internal layers is a RELU; the last layer uses the softmax function. Also, after the first layer, we added a dropout layer with 0.1 rate. 
The number of training epochs was 100. 
To identify the effect of presence or absence of features on the performance of the model, we followed two paradigms: 1) the training data includes all features listed in the Section \ref{sec:features}, 2) the training data only has the \textit{position group features}.
Also, we trained two unique models for the prediction of Unum and Index as the data labels.
Most of the previous works have employed the position features group with the "Uniform Number" sorting as their model's input to predict the UNUM of the target agent. 
Therefore, we can consider this feature group along with the UNUM as a baseline (Line 1 of Table \ref{tab:res}) for comparing the proposed features and feature groups.
Table \ref{tab:res} column \textit{Test Data}, illustrates DNN and RF's performance on the prediction of passing behavior. The obtained results show that the prediction accuracy increased when all features were present in a range of 2.03 to 4.65. Predicting the Index of pass receiver is 3.41 more accurate than predicting its UNUM. Based on the obtained results, placing the ball holder's position as the first feature of data can increase the accuracy of prediction up to 0.83 on average and up to 1.94 when the \textit{X Sorting} or \textit{Field Evaluator Sorting} is employed for sorting the data.
Finally, the \textit{UNUM Sorting} has 1.27 higher accuracy in predicting the passing behavior. 
\setlength{\abovecaptionskip}{-20pt}
\setlength{\belowcaptionskip}{-30pt}
\begin{table}[ht]
  \begin{tabular}{|l|l|l||l|l|l|l||l|l|l|l|}
    \hline
    \multicolumn{3}{|c||}{\textbf{Data set}} &
    \multicolumn{4}{c||}{\textbf{Test Data}} & 
    \multicolumn{4}{c|}{\textbf{Real Opponents}} \\
        Sorter & Kicker & Features & \multicolumn{2}{|c|}{DNN} & \multicolumn{2}{c||}{RF} & \multicolumn{2}{|c|}{DNN} & \multicolumn{2}{c|}{RF} \\
        Algorithm & First & Groups & Unum & Index & Unum & Index & Unum & Index & Unum & Index\\
      \hline
        Uniform & No & Position & 79.40 & 79.40 & 80.90 & 80.89 & 55.91 & 54.97 & 57.73 & 57.40 \\
    Uniform & Yes & Position&  79.99 & 79.94 & 81.42 & 81.43 & 55.99 & 56.43 & 58.69 & 59.19 \\
    X  & No & Position&  79.15 & 79.14 & 79.52 & 79.17 &  56.44 & 55.93 & 56.49 & 54.78 \\
    X  & Yes & Position&  79.45 & 79.32 & 80.32 & 80.38 & 57.13 & 55.68 & 58.55 & 58.93 \\
     Field Evaluator & No & Position&  79.09 & 79.15  & 79.40 & 79.17 & 57.37&55.68&55.86&54.39\\
    Field Evaluator & Yes & Position&  79.27 & 79.35  & 80.23 & 80.26 & 57.43&55.80&57.88&58.70\\
    Uniform & No & All & 84.05 & 84.15 & \textbf{84.41} & 84.10 & 63.63 & 63.29 & 62.65 & 62.69 \\
    Uniform & Yes & All&  83.74& 84.00 & 84.09 & 84.07 & 63.03 & 63.50 & 62.66 & 62.47 \\
    X  & No & All&  81.64 & 81.98 & 82.01 & 83.24 & 60.76 & 63.61 & 59.99 & 64.28 \\
    X  & Yes & All&  81.90 & 82.59 & 82.35 & 83.43 &  61.64 & 65.10 & 60.65 & 64.62 \\
    Field Evaluator & No & All&  81.55 & 82.26  & 81.92 & 83.25 &  60.24 & 64.07&59.48&63.86\\
    Field Evaluator & Yes & All&  81.97 & 82.93  & 82.33 & 83.32 &  62.17 & \textbf{65.22}&60.22&64.12\\
    
    \hline
  \end{tabular}
  \caption{The prediction accuracy of DNN and RF on the test data using different Features group and Sorting algorithms}
  \label{tab:res}
\end{table}

\begin{figure}
    \centering
    \includegraphics[scale=0.31]{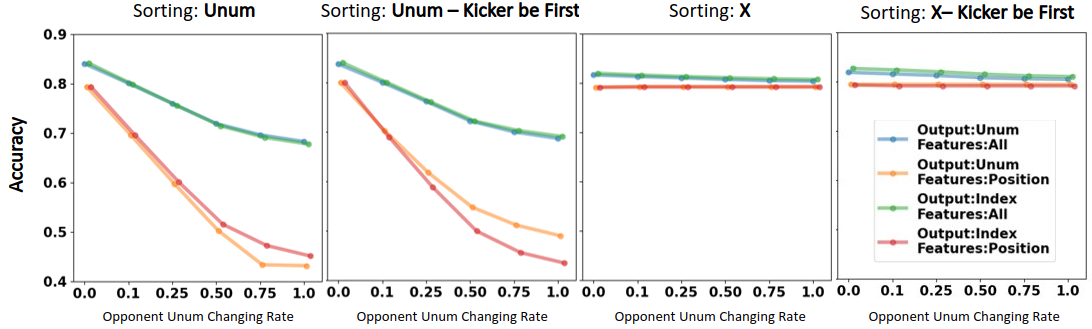} 
    \caption{The accuracy of predicting the passing behavior when different portions of opponents' UNUM were randomly changed. The percentage of players is shown in the X-Axis, and the accuracy of model on Y-Axis.}
    \label{fig:random}
\end{figure}

Soccer Simulation 2D teams sometimes change their player's uniform number. To identify robustness of pass prediction models against changing uniform number, We randomly changed (0.1, 0.25, 0.50, 0.75, 1.0) proportions of player's uniform number in the test data, and then evaluate models' accuracy for the 16 trained DNN models. Figure \ref{fig:random} presents results of this experiment. This result shows that \textit{Unum sorting} is not robust to changes of uniform number, but \textit{X sorting is robust to this change, since changing uniform number does not impact the data when the sorter algorithm is \textit{X sorting}}. Results of \textit{Field Evaluator sorting} is similar to \textit{X sorting}. It demonstrates that using all feature groups improve robustness of \textit{Uniform sorting}.

%

\vspace{-5mm}
\setlength{\abovecaptionskip}{10pt}
\setlength{\belowcaptionskip}{-30pt}

\subsection{Testing the features in games with real opponents}
Similarly to real-world soccer, in SS2D league, the rules for a player can vary between different teams; for example, in the Agent2D, uniform numbers of 2 and 3 are assigned to Center Back players, but for Cyrus, player number 2 and 3 are the Center and Right back, respectively. 
Also, the formation strategies (e.g., defensive, offensive) of the teams are generally different, and the team's pass effectiveness depends strongly on how the opponent behaves. 
In this experiment, we examine our newly created features' performance by playing games with more sophisticated opponents aiming at verifying the impacts on the passing behavior.

To evaluate the pass improvements against different opponents, we ran 60 games between the Helios Base Sample Team with Feature Extractor module against the 6 top teams of RoboCup 2019, Helios\cite{hel19}, CYRUS\cite{cyrus19}, YuShan\cite{yush19}, MT2019\cite{mt19}, Receptivity\cite{rec19} and Razi\cite{razi19}. 
After the games' execution, we generated $16,672$ data samples; then, we exploited the sorter algorithms (\ref{subsec:sorting}) to sort the new data set. 
At the end, we evaluated the accuracy of trained models with the new dataset and reported the result in Table \ref{tab:res}, column \textit{Real Opponents}.
Again, the baseline here is the first line of Table \ref{tab:res}, where we do not use the kicker first strategy and use the position features. 
The results show that the presence of all features in the data improves by almost 10\% (65.22\% against 55.91 \% of the baseline) the prediction probability of the model; also \textit{Field Evaluator Sorting} or \textit{X Sorting} has higher accuracy in the prediction of passing behavior.

The obtained results show that when all features were present the prediction accuracy increased 5.86 in average. 
Predicting the Index of pass receiver is 3.71 more accurate than predicting its UNUM when all features were present and the \textit{X Sorting} or \textit{Field Evaluator Sorting} is employed for sorting the data. Based on the obtained results, placing the ball holder's position as the first feature of data can increase the accuracy of prediction up to 1.01 in average.


\vspace{-5mm}
\subsection{Feature Importance}
Finally, to better understand the decisions of the DNN, the impact of input features on the prediction of passing strategy, we applied \textit{feature permutation} strategy. 
This strategy measures the reduction of a model's score when a single feature value has been randomly shuffled \cite{breiman2001random}. 
For the RF model, the feature importance is the normalized total reduction of the Gini criterion in the RF brought by each feature. 
We evaluated the behavior of the DNN and RF estimators on the Agent2D test data. 

\begin{figure}
    \centering
    \includegraphics[scale=0.25]{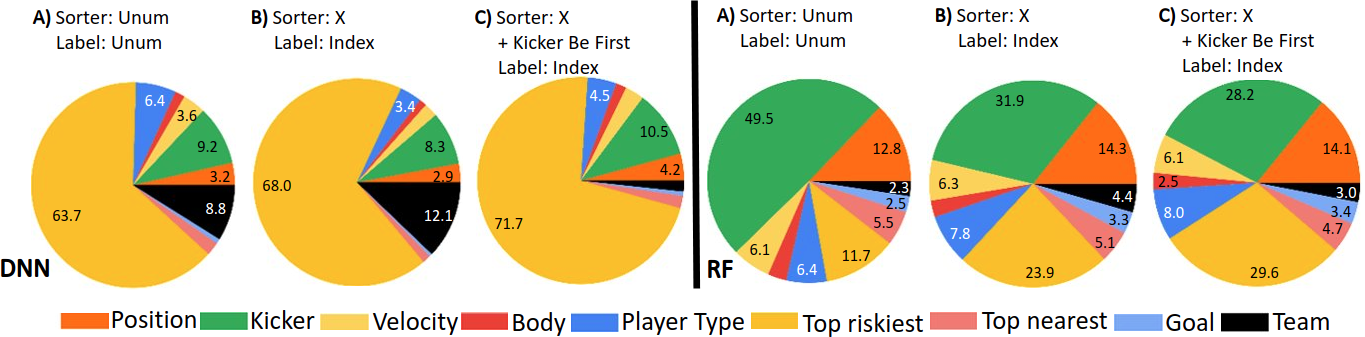} \\
    \caption{The importance of different feature group in the presence of different sorting algorithms}
    \label{fig:import1}
\end{figure}

The results of \textit{Feature Permutation} for the trained DNN models and the features importance for RF are shown in Figure \ref{fig:import1}. 
The explained models were structured as follows:
(a) input data were sorted by UNUM, and the label was the UNUM of a player;
(b) input data were sorted by \textit{X Position}, and the label was the Index of a player;
(c) input data were sorted by \textit{X Position}, the \textit{Kicker First} was $True$, and the label was the Index of player.

The result of DNN shows that the \textit{Top Riskiest opponents} feature group has the greatest influence on the prediction probability of passing behavior among all of the models, as can be seen in Figure \ref{fig:import1}. 
After the top-riskiest opponents, the \textit{Kicker's position} has the next highest impact on the prediction of the DNN, mainly when the sorting algorithm uses the ball holder position player as the first element of the data. RF shows relatively similar behavior, in which the \textit{kicker's position} and \textit{top Riskiest} features groups have the most impact on the prediction of the model. Surprisingly for both of these models, the \textit{Position} features have relatively low importance. From this chart, we can observe that although the process of learning is different for these models, similar results can be achieved when our newly engineered features are added.

\begin{figure}
    \centering
    \includegraphics[scale=0.39]{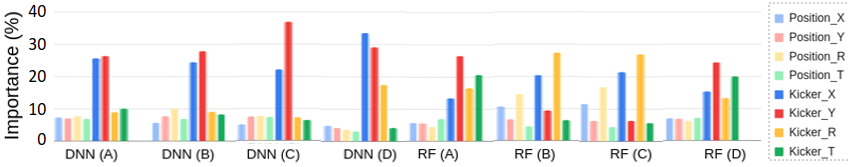} \\
    \caption{The impact of "Kicker" feature group on the prediction of DNN and RF in the presence of other position related features}
    \label{fig:import2}
\end{figure}

Most previous works on this topic have exploited the \textit{Position} features group of elements in the field. 
In this part, we want to observe the impact of \textit{Kicker} features group on the prediction of DNN and RF even in the presence of \textit{Position} features.   
Hence, we have compared the importance of the \textit{Kicker} feature group regarding the \textit{Position} features for 5 models. The obtained results are illustrated in Figure \ref{fig:import2}. 

In models A - C, all groups of features were used for the predictor model. In all of them, the \textit{Kicker} features group has a higher impact on the prediction of the DNN and RF than \textit{Position} features. 
The impact of \textit{Kicker} features is more evident for model D when only two feature groups were used for this model. 
For model D, we employed both of \textit{position} and \textit{Kicker} features groups to examine the impact of positional features group on the prediction of the model.

Fig \ref{fig:import2} shows the result of our experiment. The results show that the \textit{Kicker} group is more important than the \textit{Position} group; specially for DNN, $X$ and $Y$ of the \textit{Kicker} group are the most important features in these two groups.

Since the \textit{Kicker} feature group measures each element's position with respect to the ball holder, the location of objects is measured based on the ball's location. Using this feature group, our model can consider the similar situation in different locations of the field, and in this way becoming translation invariant. 
Additionally, similar to a real soccer game, if the players (teammates and opponents) are closer to the ball holder, the chance of receiving the ball or losing the ball is higher; hence the position of the objects near the kicker is crucial.
\vspace{-5mm}
\setlength{\belowcaptionskip}{-5pt}
\vspace{-10mm}
\vspace{-10mm}
\section{Conclusion}
\label{sec:conclusion}

In this paper, we proposed an embedded data generator module in the Agent 2D base to record the information of a game SS2D game when the player is the ball holder aiming to create features to be used by machine learning algorithms to predict the best option for passing. 
To model our teammates' passing behavior in the game, we exploited the generated features and trained a Deep Neural Network and a Random Forest. 
We obtained an 84\% accuracy on the prediction of passing behavior on the test data. 
Furthermore, we tested the trained models in terms of robustness against changes in the opponent's team, prediction of the accuracy facing the top five teams of RoboCup 2019, and the effectiveness of different feature groups on the prediction of passing behavior.

Our experiment shows that "X" sorting methods enhanced the models to predict agents' passing targets against varieties of opponents and situations.
Also, the comparison of different features groups showed that features related to the position of the ball holder have higher importance for the prediction of pass target than other positional features.

We intend to explore several other directions in this work. First, we will disable the full state mode in the soccer simulation server to receive noisy information from the server and verify how the new features would impact in the result of games (i.e., win, tie, loss).
We also intent to use other models with the newly engineered features such as recurrent neural networks. 


\begin{thebibliography}{9}






\bibitem{mackworth1993seeing}
A. K. Mackworth. “On seeing robots”. In:Computer Vision: Systems, Theory and Applications. World Scientific, 1993, pp. 1–13.
\bibitem{nakashima2015kick}
Nakashima, T., Mifune, S., Henrio, J., Obst, O., Wang, P., Prokopenko, M.: Kick extraction for reducing uncertainty in RoboCup logs. In:International Conference on Human Interface and the Management of Information. Springer. 2015,pp. 622–633.

\bibitem{fukushima2019similarity}
Fukushima, T., Nakashima, T., Akiyama. H.: Similarity analysis of action trajectories based on kick distributions. In:Robot World Cup. Springer. 2019, pp. 58–70.

\bibitem{michael2017analysing}
Michael, O., Obst, O., Schmidsberger, F., Stolzenburg, F.: Analysing soccer games with clustering and conceptors. In:Robot WorldCup. Springer. 2017, pp. 120–131.

\bibitem{asali2016using}
Asali, E., Valipour, M., Zare, N., Afshar, A., Katebzadeh, M., Dastghaibyfard, GH.: Using Machine Learning approaches to detect opponent formation. In:2016Artificial Intelligence and Robotics (IRANOPEN). IEEE. 2016, pp. 140–144.

\bibitem{suzuki2019use}
Suzuki, Y., Nakashima, T.: On the use of simulated future information for evaluating game situations. In:Robot World Cup. Springer. 2019, pp. 294–308.

\bibitem{kitano1997robocup}
Kitano, H., Asada, M., Kuniyoshi, Y., Noda, I., Osawa, E., Matsubara, H.: RoboCup: A challenge problem for AI. In:AI magazine18.1 (1997), pp. 73–73.

\bibitem{akiyama2013helios}
Akiyama, H., Nakashima, T.: Helios base: An open source package for the robocup soccer2d simulation. In:Robot Soccer World Cup. Springer. 2013, pp. 528–535.

\bibitem{prokopenko2019gliders2d}
Prokopenko, P., Wang, P.: Gliders2d: source code base for RoboCup 2D soccer simulation league. In:Robot World Cup. Springer. 2019, pp. 418–428.

\bibitem{kok2003uva}
Kok, J., Vlassis, N., Groen, F.: UvA Trilearn 2003 team description. In:Proceedings CD RoboCup 2003 (2003).



\bibitem{mugnier1992conceptual}
Mugnier, M.-L., Chein, M.: Conceptual graphs: Fundamental notions. In:Revue d’intelligence artificielle 6.4 (1992), pp. 365–406.

\bibitem{hel19}
Akiyama, H., Nakashima, T., Fukushima,T., Suzuki, Y., Ohori, A.: HELIOS2019: Team Description Paper. In:RoboCup(2019).
 
\bibitem{cyrus19}
Zare, N., Sarvmaili, M., Mehrabian, O., Nikanjam, A., Khasteh, S.-H., Sayareh, A., Amini, O., Barahimi, B., Majidi, A., Mostajeran, A.: Cyrus 2D Simulation 2019. In:RoboCup(2019).
 
\bibitem{yush19}
Cheng, Z., Xie, N., Zhang, F., Guang, B., Zhang, Q., Sun, C., Wang, X., Wang, L.: YuShan Team Description Paper for RoboCup2019. In:RoboCup(2019).

\bibitem{mt19}
Wang, X., Dong, N., Li, Ch., Zhang, X., Liu, G., Fang, Y., Chen, Sh., Hu, Ch.: MT2019 Robocup Simulation 2D Team Description. In:RoboCup(2019).

\bibitem{rec19}
Li, M.: Receptivity: Team Description Paper 2018 Fine Tuning of Agent Decision Evaluation. In:RoboCup(2019).

\bibitem{razi19}
Noohpisheh, M., Shekarriz, M., Bordbar, A., Liaghat, M., Salimi, A., Borzoo, D., Zarei, A.: Razi Soccer 2D Simulation Team Description Paper 2019. In:RoboCup(2019).

\bibitem{breiman2001random}
Breiman, L.: Random forests. In:Machine learning45.1 (2001), pp. 5–32.

\end{thebibliography}
\end{document}